\documentclass[letterpaper, 10 pt, conference]{ieeeconf}  % Comment this line out if you need a4paper

\IEEEoverridecommandlockouts                              % This command is only needed if 
\usepackage{cite}
\usepackage{amsmath,amssymb,amsfonts}
\usepackage{algorithm}
\usepackage{algpseudocode}
\usepackage{graphicx}
\usepackage{textcomp}
\usepackage{xcolor}
\usepackage{subfigure}
\def\BibTeX{{\rm B\kern-.05em{\sc i\kern-.025em b}\kern-.08em
    T\kern-.1667em\lower.7ex\hbox{E}\kern-.125emX}}
\usepackage{theorem}

\newtheorem{proposition}{Proposition}

\usepackage{geometry}
\title{\LARGE \bf
Bayesian Generalized Kernel Inference for Exploration \\of Autonomous Robots
}

\author{Yang Xu, \IEEEmembership{Student Member, IEEE,} Ronghao Zheng$^{\dagger}$, \IEEEmembership{Member, IEEE,}
        Senlin Zhang, \IEEEmembership{Member, IEEE,}\\
        and Meiqin Liu, \IEEEmembership{Senior Member, IEEE}% <-this % stops a space
\thanks{$^1$Yang Xu, Ronghao Zheng and Senlin Zhang are with the College of Electrical Engineering, Zhejiang University, Hangzhou 310027, China. 
		\texttt{\{xuyang94,rzheng,slzhang\}@zju.edu.cn}}
\thanks{$^2$Meiqin Liu is with the Institute of Artificial Intelligence and Robotics, Xi'an Jiaotong University, Xi'an 710049, China. \texttt{liumeiqin@zju.edu.cn}}
\thanks{$^3$All authors are also with the State Key Laboratory of Industrial Control Technology, Zhejiang University, Hangzhou 310027, China.}
\thanks{$^{\dagger}$Corresponding author}
}

\begin{document}

\maketitle
\thispagestyle{empty}
\pagestyle{empty}

%%%%%%%%%%%%%%%%%%%%%%%%%%%%%%%%%%%%%%%%%%%%%%%%%%%%%%%%%%%%%%%%%%%%%%%%%%%%%%%%
\begin{abstract}

This paper concerns realizing highly efficient information-theoretic robot exploration with desired performance in complex scenes. 
We build a continuous lightweight inference model to predict the mutual information (MI) and the associated prediction confidence of the robot's candidate actions which have not been evaluated explicitly. 
This allows the decision-making stage in robot exploration to run with a logarithmic complexity approximately, this will also benefit online exploration in large unstructured, and cluttered places that need more spatial samples to assess and decide.
We also develop an objective function to balance the local optimal action with highest MI value and the global choice with high prediction variance. 
Extensive numerical and dataset simulations show the desired efficiency of our proposed method without losing exploration performance in different environments. 
We also provide our open-source implementation codes released on GitHub for the robot community.
\end{abstract}

%%%%%%%%%%%%%%%%%%%%%%%%%%%%%%%%%%%%%%%%%%%%%%%%%%%%%%%%%%%%%%%%%%%%%%%%%%%%%%%%
\section{Introduction}

\begin{figure}
	\centering
% 	\vspace{0.3em}
	\subfigure[]{
		\begin{minipage}[t]{0.5\linewidth}
			\centering
			\includegraphics[width=1.8in]{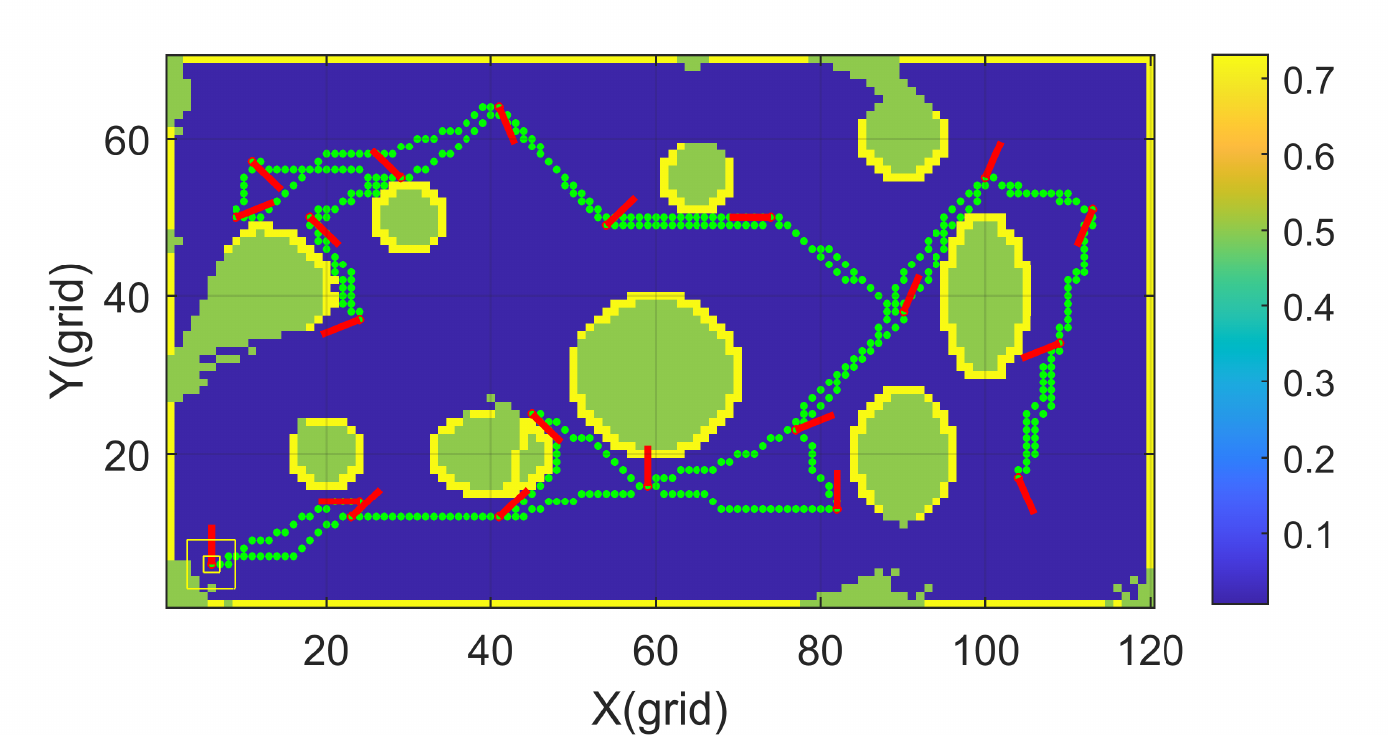}
			%\caption{fig1}
		\end{minipage}%
	}%
% 	\quad
	\subfigure[]{
		\begin{minipage}[t]{0.5\linewidth}
			\centering
			\includegraphics[width=1.8in]{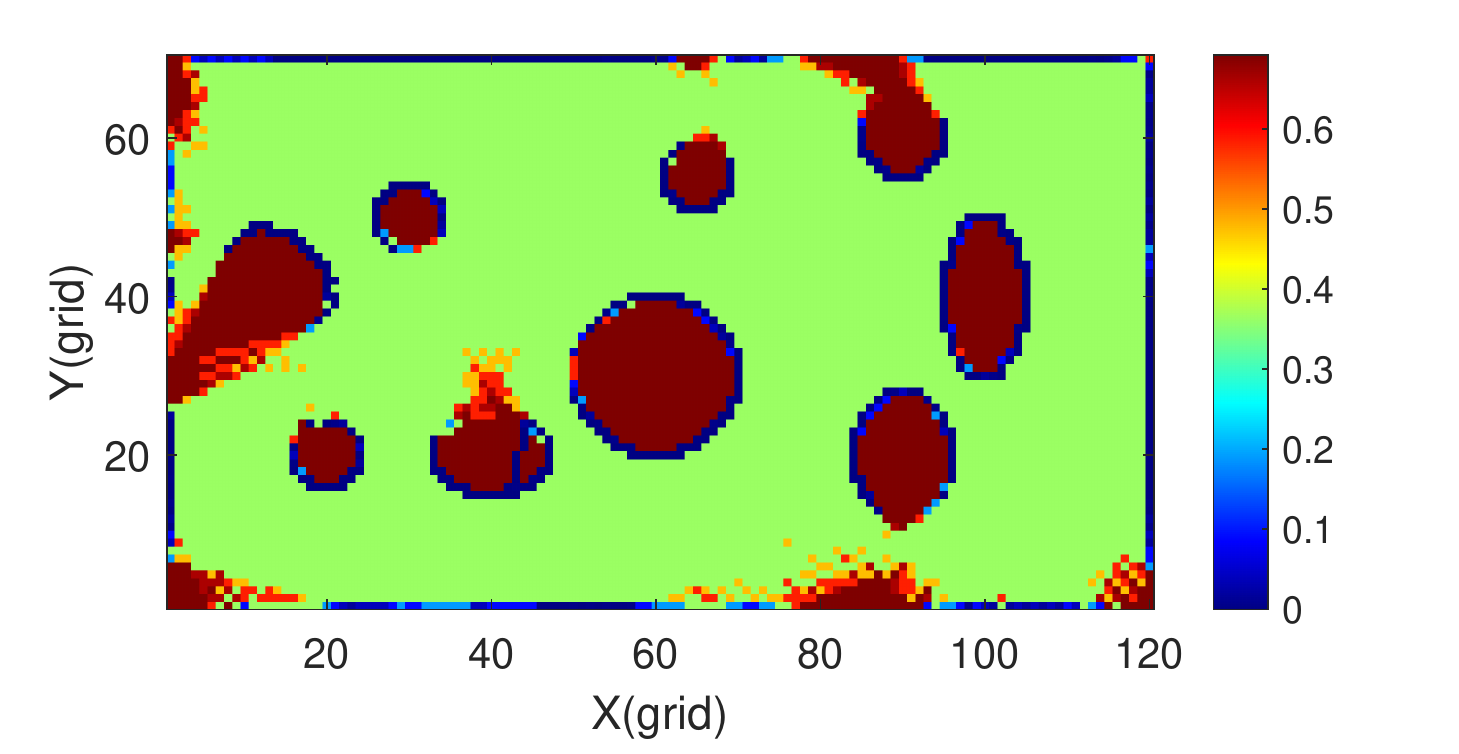}
			%\caption{fig2}
		\end{minipage}%
	}%
% 	\vspace{0.3em}
	\centering
	\caption{MI-based active robot exploration in an unknown \textbf{unstructured} environment. (a) Informative trajectory and the resulting occupancy map, (b) Resulting MI surface. Note that the coincident yellow squares mean the start and end points. A minimum information threshold is set to select more informative exploration actions, e.g. the middle left and top right areas are less informative than the threshold and thus unexplored. Note that the scale of MI is in [0,1] bit in this paper.}
	\label{fig:unstr-traj}
\end{figure}

Robot exploration gains its prevalence recently in \textit{priori} unknown environments such as subterranean, marine, and planetary tasks
\cite{Azpurua2021three,strader2020perception,stankiewicz2021adaptive}.
Among the literature, state-of-the-art exploration methods prefer to use information-theoretic metrics in each iteration, such as Shannon mutual information (MI) \cite{julian2014mutual} and its derivatives \cite{charrow2015information,zhang2020fsmi,yang2021crmi,xu2022confidence}, to evaluate the information gain brought by candidate control actions accurately and choose and execute the most informative action, thus the exploration problem becomes a sequential optimal decision-making one naturally. A typical exploration example is in Fig.~\ref{fig:unstr-traj}.

Intuitively, the way to tackle this problem is to use a greedy strategy and add more candidate actions, including sampled nodes \cite{hollinger2014sampling,ghaffari2019sampling}, available viewpoints \cite{bircher2016receding, charrow2014approximate}, or special motion primitives \cite{yang2013gaussian,charrow2015rss}, in the discrete action space. 
However, the exploration performance of greedy selection is closely related to the discrete sampling resolution/method of action space over the map grid, i.e., a coarse resolution may lead to sub-optimal actions/paths, and a fine one may generate more samples and be more likely to choose the optimal action, but the computational cost of the information gain evaluation of all candidate actions will become expensive in this case since the forward simulation in the evaluation requires extensive raycasting and MI calculation. Notably, these consequences will be more distinct in 3D environments because the increased dimension needs much more samples.

In this paper, we aim to realize a more efficient and accurate approach to find the most informative action without evaluating all candidate actions exhaustively and expensively in robot exploration. 
Specifically, our main contributions are three-fold: 

1) We propose a Bayesian kernel spatial MI inference method to construct a continuous surrogate evaluation model between robot actions and MI values using only partial explicitly evaluated samples, which can perform highly efficient MI prediction of control actions in logarithm time;

2) We develop a reward function comprising the predicted MI values and uncertainties to find the best action for realizing the trade-off between exploration and exploitation, which has been validated in numerical and dataset simulations;

3) Meanwhile, we release an open-source implementation of our proposed method here\footnote{https://github.com/Shepherd-Gregory/BKI-exploration} for the robotics community.

The paper organization is as follows. Related works about the recent learning-based robot exploration methods are presented in Section II. We formulate the problem in Section III and present our Bayesian kernel-based MI inference method in Section IV. Simulation results using synthetic data and the real world dataset and discussions are given in Section V, followed by conclusions in Section VI. 
%%%%%%%%%%%%%%%%%%%%%%%%%%%%%%%%%%%%%%%%%%%%%%%%%%%%%%%%%%%%%
%%%%%%%%%%%%%%%%%%%%%%%%%%%%%%%%%%%%%%%%%%%%%%%%%%%%%%%%%%%%%

\section{Related Work}

In the context of robot exploration, supervised learning techniques provide a powerful tool to find the global optimum approximately by training predictive models using minor parts of actions in continuous action spaces, without evaluating the objective function expensively, which also has better interpretability in black-box inference\cite{marchant2014bayesian,oliveira2020bayesian,francis2019occupancy}. 

In \cite{bai2015inference}, Bai \textit{et al.} used the Gaussian process (GP) to model the relationship between control actions and the explicitly evaluated MI for the robot exploring \textit{priori} unknown areas.
In \cite{bai2016information}, they further introduced Bayesian optimization (BO) into the information-theoretic robot exploration to optimize the GP prediction in multiple iterations, which provides rapid map entropy reduction and ensures computational efficiency. 
Generally, BO assumes a prior distribution on the objective function and constructs predictive models to describe the underlying relationship between robot actions and their MI.
It also assesses the acquisition function derived from the GP prior and samples, then chooses the next query point maximizing the acquisition function and balancing the trade-off between exploration (global) and exploitation (local). Iteratively, BO presents more precise results on the posterior distribution as the observations (training samples) increase.
Rather than evaluating discrete viewpoints, Francis \textit{et al.}\cite{francis2019occupancy} modeled the autonomous exploration and mapping task as a constrained BO aiming to find optimal continuous paths.

However, the main bottleneck of the above BO-based robot exploration methods is that the number of the training actions $N$ will affect the resulting prediction accuracy directly, as well as the computational cost. That implies one needs to pay expensive computations to achieve higher exploration performance. Typically, updating and querying the GP models (the engine behind BO) have an overall $\mathcal{O}(N^3)$ time complexity. This compromises the inference efficiency and real-time performance of robot exploration tasks inevitably, especially in large-scale and 3D scenes.

More recently, deep neural networks (DNNs) have been introduced to realize predicting optimal sensing actions more efficiently. Bai \textit{et al.} \cite{bai2017toward} trained the DNN with plenty of randomly generated 2D maps to generate suggested action and ensure inferring in constant time. Graph neural networks (GNNs) have also been combined with reinforcement learning methods to learn the best action from an exploration graph, rather than metric maps or visual images
\cite{chen2020autonomous,wang2018nervenet}. 
Nevertheless, the neural network-based robot exploration methods require numerous training samples beforehand and are also limited to the adaptability and generalization capability in different environments, which may need further studies in the future.

Encouragingly, the Bayesian kernel inference (BKI) technique proposed in \cite{vega2014nonparametric} gives us a chance to perform efficient exact inference on a simplified model, rather than approximating inference on an exact generative model (e.g. GP) expensively. BKI extends local kernel estimation to Bayesian inference for exponential family likelihood functions, enabling only $\mathcal{O}(\log N_q)$ ($N_q$: the number of querying samples) run time for inference. These significant merits enhance BKI's application in robotics, including sensor uncertainty estimation \cite{valentin2016predictive}, high-speed navigation \cite{richter2015bayesian}, as well as environment mapping using sparse sensor measurements such as terrain traversability mapping \cite{shan2018bayesian}, 3D occupancy mapping \cite{doherty2019learning}, semantic mapping\cite{lu2020bayesian}. 

Motivated by \cite{bai2016information} and \cite{vega2014nonparametric}, use BKI to infer the spatial MI in an efficient and closed-form way for the control actions whose MI values have not been explicitly evaluated via expensive computation (e.g. \cite{julian2014mutual}). Our method keeps similar accuracy to previous approaches compared with existing works such as \cite{bai2015inference} and \cite{bai2016information}, but shows more efficient and suitable performance for complex scenes requiring numerous explicitly evaluated samples.
\section{Preliminaries and Notions}

In this paper, for simplicity of discussion, we mainly consider the information-theoretic exploration using a mobile robot equipped with a beam-based range sensor of limited field of view (FOV) in a 2D environment. The results here can also be extended to 3D cases expediently. 

\subsection{Information-Theoretic Exploration}
Generally, the robot generates a set of candidate actions  $\mathcal{X}_{action}$ in the robot's feasible configuration space $\mathcal{X}\subseteq SE(2)$. We also assume this configuration space has been discretized by a fixed resolution over the 2D static grid map. The set of values $m\in [0,1]$ is the occupancy level over the independent grid cells and can be updated and queried by the classic log-odds method \cite{thrun2005probabilistic}. The occupancy value of an unobserved map cell $\xi$ is assumed to be uniform, i.e., $p(m_\xi) = 0.5$. 

Here we use the classic Shannon MI \cite{julian2014mutual} as the information measure of candidate configuration $x_i=[p_i^x, p^y_i, \psi_i] \in \mathcal{X}_{action}$, where $p_i^x$ and $p^y_i$ denote the robot's position on the map, and $\psi_i$ denotes the heading angle of the robot. From the view of information theory, the expected information gain of $x_i$ can be evaluated by the current map entropy and conditional entropy given a new measurement at $x_i$:
\begin{equation}
    I(m;x_i) = H(m) - H(m|x_i).
    \label{eq:mi}
\end{equation}

The aim of information-theoretic robot exploration is to select the best action $x_{best}$ maximizing the expected MI:
\begin{equation}
    x_{best}=\mathop{\mathrm{argmax}}\limits_{x\in \mathcal{X}_{action}} I(m;x_i).
    \label{eq:maxmi}
\end{equation}

Notably, the MI of each configuration can be decomposed over independent beams and then to cells via raycasting, then accumulated MI over cells to approximate, which owns a squared time complexity in map resolution $\lambda_m$ at worst \cite{yang2021crmi}. This also brings more evaluation costs for robot exploration.

\subsection{Bayesian Generalized Kernel Inference}
Consider a supervised learning-based inference problem on predictive stochastic models $p(\mathbf{y}|\mathbf{x})$ given a sequence of $N$ observations $\mathcal{D}=\{(\mathbf{x}=\{x_i\},\mathbf{y}=\{y_i\})\}_{i=1}^N$, where $\mathbf{x}$ and $\mathbf{y}$ represent the set of evaluated configurations and the resulting MI values $I(m;\mathbf{x})$, respectively. The main objective is to infer the posterior distribution $p(\mathbf{y}^*|\mathbf{x}^*,\mathcal{D})$ for the target inputs $\mathbf{x}^*$ to be evaluated. 
This problem can be solved by associating latent parameters $\boldsymbol \theta=\{\theta_i\}_{i=1}^N \in \mathbf \Theta$ with input $\mathbf{x}$ in the latent space $\mathbf \Theta$, where the likelihood $p(\mathbf{y}|\boldsymbol{\theta})$ is known. Thus the inference on $\mathbf{y}^*$ can be formulated as an inference on target parameters $\boldsymbol{\theta}^*$ related to $\mathbf{x}^*$:
\begin{equation}
	p(\mathbf{y}^*|\mathbf{x}^*,\mathcal{D}) = \int_{\mathbf \Theta} p(\mathbf{y}^*|\boldsymbol{\theta}^*) p(\boldsymbol{\theta}^*|\mathbf{x}^*,\mathcal{D}) d\mathbf{\theta}^*,
	\label{eq:post}
\end{equation}
where the posterior distribution of the latent variables can be characterized using Bayes' rule:
$
	p(\boldsymbol{\theta}^*|\mathbf{x}^*,\mathcal{D}) \propto \int_{\mathbf \Theta} \prod_{i=1}^{N} p(y_i|{\theta}_i)p({\theta}_{1:N},\boldsymbol{\theta}^*|x_{1:N},\mathbf{x}^*)d {\theta}_{1:N}.
$

By strongly assuming latent parameters ${\theta}_{1:N}$ are conditionally independent given the target parameters $\mathbf{\theta}^*$:
$
	p({\theta}_{1:N},\boldsymbol{\theta}^*|a_{1:N},\mathbf{x}^*) = \prod_{i=1}^{N} p(\theta_i|\boldsymbol{\theta}^*,x_i,\mathbf{x}^*) p(\boldsymbol{\theta}^*|\mathbf{x}^*)
$, one can marginalize the latent variables ${\theta}_{1:N}$ and then obtain 
$
	p(\boldsymbol{\theta}^*|\mathbf{x}^*,\mathcal{D}) \propto \prod_{i=1}^{N} p(y_i|\boldsymbol{\theta}^*,x_i,\mathbf{x}^*) p(\boldsymbol{\theta}^*|\mathbf{x}^*).
$

BKI further defines a distribution that has a special smoothness constraint and bounded Kullback-Leibler divergence (KLD) $D_{KL}(g||f)$ between the extended likelihood $p(y_i|\boldsymbol{\theta}^*,x_i,\mathbf{x}^*)$ represented by $g$ and the likelihood $p(y_i|\theta_i)$ represented by $f$, i.e., the maximum entropy distribution $g$ satisfying $D_{KL}(g||f)\leq \rho(\mathbf{x}^*,\mathbf{x})$ has the form $g(\mathbf{y})\propto f(\mathbf{y})^{k(\mathbf{x}^*,\mathbf{x})}$, where $\rho(\cdot,\cdot):\mathcal{X}\times \mathcal{X}\to \mathbb{R}^+$ is a smoothness bound and $k(\cdot,\cdot):\mathcal{X}\times \mathcal{X}\to [0,1]$ is a kernel function which can be uniquely determined by $\rho$. 
Substituting into Eq.~\eqref{eq:post}, we can get:
\begin{equation}
 p(\boldsymbol{\theta}^*|\mathbf{x}^*,\mathcal{D}) \propto \prod_{i=1}^{N} p(y_i|\boldsymbol{\theta}^*)^{k(\mathbf{x}^*,\mathbf{x})} p(\boldsymbol{\theta}^*|\mathbf{x}^*)
 \label{eq:exp}
\end{equation}

Thus the posterior distribution can be exactly inferred by using the likelihood from the exponential family and assuming the corresponding conjugate prior.

\section{Bayesian Kernel Inference for Robot Exploration}
To efficiently evaluate the exact MI of unknown robot configurations sampled in the spatial action space, we solve this problem by a Bayesian kernel inference way.
\subsection{Bayesian Kernel Spatial MI Inference}
As mentioned in Section III.B, we assume the underlying likelihood model between the MI values $\mathbf{y}$ and the latent parameters $\boldsymbol{\theta}$ follows Gaussian distribution with unknown mean $\boldsymbol{\mu}\in \mathbb{R}^N$ and fixed, known covariance $\boldsymbol{\Sigma}$:
\begin{equation}
	p(\mathbf{y}|\boldsymbol{\mu}) = \mathcal{N} (\boldsymbol{\mu}, \boldsymbol{\Sigma}), \boldsymbol{\Sigma}=diag(\sigma^2)\in \mathbb{R}^{N\times N},
	\label{eq:muy}
\end{equation}
thus its conjugate prior can also be described by a Gaussian distribution using the hyperparameter $\zeta$ and target samples input $\mathbf{x}^*$:
\begin{equation}
	p(\boldsymbol{\mu}|\mathbf{x}^*) = \mathcal{N} \left(\boldsymbol{\mu}_0(\mathbf{x}^*), \frac{1}{\zeta(\mathbf{x}^*)}\boldsymbol{\Sigma}(\mathbf{x}^*)\right),
	\label{eq:mux}
\end{equation}
where $\boldsymbol{\mu}_0$ and $\zeta$ are the initial belief of the mean and the uncertainty of that belief, respectively. $\zeta=0$ means no confidence and $\zeta\to \infty$ indicates full prior knowledge. 
Here we assume $\zeta$ is a quite small positive constant since we do not have much prior information about the belief when exploring unknown areas.
 
Therefore, we can substitute Eq.~\eqref{eq:mux} and Eq.~\eqref{eq:muy} into Eq.~\eqref{eq:exp} given observations $\mathcal{D}$:
\begin{eqnarray}
	p(\boldsymbol{\mu}^*|\mathbf{x}^*,\mathcal{D}) \propto \prod_{i=1}^{N} \exp \left(-\frac{1}{2} \frac{(y_i-\mu_i)^2}{\sigma^2} k(\mathbf{x}^*,x_i)\right) \\ \nonumber
	\cdot \exp \left(-\frac{1}{2} \frac{(\mu_i-\mu_0)^2}{\sigma^2} \zeta \right),
\end{eqnarray}
and the posterior over mean and covariance of the MI can be derived as follows:
\begin{equation}
	\begin{split}
	 I(\mathbf x^*)=\mathbb{E}[\mathbf y^*|\mathbf{x}^*,\mathcal{D}]=\mathbb{E}[\boldsymbol{\mu}^*|\mathbf{x}^*,\mathcal{D}]=\frac{\overline{\mathbf{y}}+\zeta\mu_0}{\zeta+\overline k}\simeq \frac{\overline{\mathbf{y}}}{\overline{k}}, \\ 
	 \sigma_I(\mathbf x^*)=\mathbb{V}[\boldsymbol{\mu}^*|\mathbf{x}^*,\mathcal{D}]=\frac{\boldsymbol{\Sigma}}{\zeta+\overline k} \simeq \frac{\boldsymbol{\Sigma}}{\overline k}, 
	\label{eq:mivar}
	\end{split}
\end{equation}
where $\overline{\mathbf{y}}$ and $\overline{k}$ can be computed by kernel functions:
\begin{equation}
	\overline{k} = \Sigma_{i=1}^N k(\mathbf{x}^*,\mathbf{x}),~
	\overline{\mathbf{y}} = \Sigma_{i=1}^N k(\mathbf{x}^*,\mathbf{x}) y_i.
	\label{eq:barky}
\end{equation}

Give a set of observations $\mathcal{D}$ evaluated explicitly as the input, then we can easily compute the MI and the corresponding confidence for the test spatial configurations $\mathbf{x}^*$ by using Eq.~\eqref{eq:mivar} and Eq.~\eqref{eq:barky}. 

\subsection{Kernel Selection}
The kernel function of the BKI method will directly affect the computational efficiency and accuracy, thus selecting an appropriate kernel is quite significant. In \cite{shan2018bayesian,doherty2019learning,lu2020bayesian}, the chosen sparse kernels remove the training points far away from the queried points, which allows efficient and exact evaluation (e.g. occupancy, traversability, semantic class) over the observations in logarithm run time using k-d trees.

Unlike the sufficient training data obtained from onboard sensors in mapping tasks, robot exploration always generates and evaluates relatively fewer candidate configurations in a limited space at each time instance, so there is no need to reject the rare training samples in robot exploration tasks.

Among the exponential kernel functions, we prefer the Mat$\acute{\text e}$rn kernel for its capability of handling sudden transitions of terrain \cite{rasmussen2005gaussian,ramos2012gaussian}, since the potential obstacles and unknown structures in application scenes that have never been seen before will vary the MI values greatly. The typical Mat$\acute{\text e}$rn kernel function is as follows:
\begin{equation}
	k(\mathbf{x}^*,\mathbf{x}) = \frac{2^{1-\nu}}{\Gamma(\nu)}(\frac{\sqrt{2\nu}r}{\ell})^{\nu}K_{\nu}(\frac{\sqrt{2\nu}r}{\ell}), r=||\mathbf{x}^*-\mathbf{x}||,
\end{equation}
where the positive parameters $\nu$ and $\ell$ are the smoothness constant and characteristic length scale respectively, $\Gamma(\cdot)$ and $K_{\nu}$ are the gamma and modified Bassel function, respectively.
In practice, we choose a Mat$\acute{\text e}$rn 3/2 kernel ($\nu=3/2$) with the form as 
$k(\mathbf{x}^*,\mathbf{x}) = (1+\frac{\sqrt{3}r}{\ell})\exp (-\frac{\sqrt{3}r}{\ell})$.

\subsection{BKI-based Robot Exploration}
In robot exploration, we expect the robot moves toward the places with high predicted MI values to maximize the information gain locally, but this greedy ``exploration'' may lead to undesired paths or even worse such as getting stuck in cluttered areas. Instead, the unexplored places with high predicted uncertainty are also worth exploring, since they may guide an optimal path for the robot globally in a prior unknown area, which is also characterized as ``exploitation''.

Therefore, we integrate the prediction confidence of MI values with the predicted MI to realize a trade-off between the exploration and exploitation, then we can get the suggested action maximizing the information objective function based on Eq.~\eqref{eq:maxmi} and Eq.~\eqref{eq:mivar}:
\begin{equation}
	x_s=\mathop{\mathrm{argmax}}\limits_{x\in \mathcal{X}_{action}} \alpha I(m;x_i)+ (1-\alpha)\sigma_I(x_i),
	\label{eq:ucb}
\end{equation}
where $\alpha\in [0,1]$ is the trade-off factor.

The autonomous exploration framework based on our BKI MI inference method is given in Algorithm \ref{alg:bkiexp}, where Algorithm \ref{alg:bkiopt} is the BKI optimization module.

\begin{algorithm}[ht] 
	\caption{BKI Exploration(~)} 
	\begin{algorithmic}[1]
		\Require {Occupancy map at $k$th time step $m_k$, previous robot poses $x_{hist}=x_{0:k-1}$ and current pose $x_k$, the number of explicit evaluated samples $N$, information threshold $I_{th}$, the number of querying samples $N_{q}$, while-loop counts limit $N_{loop}$}
		\State {iter = 0}
		\While {$x_{hist} \neq \emptyset$ AND iter $<N_{loop}$}
		\State {iter = iter + 1}
		\State //~\textit{Sample $N$ training actions}
		\State {$\mathbf x \leftarrow Sampling(x_k, m_k, N)$;}
		\State //~\textit{Evaluate these actions explicitly} Eq.~\eqref{eq:mi}
		\For {each $x_i\in \mathbf x$}
		\State {$m_{virtual}\leftarrow Raycasting(x_i,m_k)$;}
		\State {$I_i \leftarrow ComputeMI(m_{virtual})$;}
		\State {$\mathbf y \leftarrow \mathbf y \cup I_i$;}
		\EndFor
		\State {$\mathbf x^* \leftarrow Sampling(x_k, m_k, N_{q})$;}
		\State //~\textit{Find the suggested action using Algorithm~\ref{alg:bkiopt}}
		\State {$\{x_{best}, I_{best}\} \leftarrow BKIOptimization(\{\mathbf x,\mathbf y\},\mathbf x^*)$;}
		\If{$max(I_{best})>I_{th}$}
		\State {$x_{k+1} \leftarrow x_{best}$(MaxInfoIndex);}
		\State {$x_{hist} \leftarrow x_{hist} \cup x_{k+1}$;}
		\Else
		\State {$x_{k+1} \leftarrow x_{k-1};$ // \textit{Back to previous action}}
		\State Remove $x_{k-1}$ from $x_{hist}$;
		\EndIf
		\State // \textit{Execute the action and update the map}
		\State $P_{local}\leftarrow Astar(x_k,x_{k+1})$ // \textit{Plan local path by A*}
		\State {$m_{k+1} \leftarrow OccupancyGridMapping(P_{local})$;}
		\EndWhile 
	\end{algorithmic} 
	\label{alg:bkiexp}
\end{algorithm}

\begin{algorithm}[ht] 
	\caption{BKI Optimization(~)} 
	\begin{algorithmic}[1]
		\Require {Training set $\mathcal{D}=\{(x_i, y_i)\}_{i=1}^N$, current action set to be evaluated $\mathbf x^*$, training epoch $N_{epoch}$, factor $\alpha$}
		\State {$x_{best} \leftarrow \{\}, I_{best} \leftarrow \{\}$;}
		\For {each epoch}
		\State //~\textit{Compute the kernel function using} Eq.~(11)
		\State {$k \leftarrow KernelFunction(\mathbf x^*,\mathbf x)$;}
		\State //~\textit{Compute MI and uncertainty using} Eq.~\eqref{eq:mivar}		
		\State $\overline{k} \leftarrow \Sigma k,~\overline{y} \leftarrow k\cdot\mathbf y$;
		\State $I^* \leftarrow \overline{y}/\overline{k},~\sigma_I^* \leftarrow \Sigma/\overline{k}$;
		\State $ObjFunc \leftarrow \alpha I^*+ (1-\alpha)\sigma_I^*$;
		\State $x_s = max(ObjFunc)$;
		\If{$x_s \in \mathbf x$}
		\State $x_{best} \leftarrow x_{best} \cup x_s, I_{best} \leftarrow I_{best} \cup y_s$
		\Else 
		\State // \textit{Evaluate MI explicitly using} Eq.~\eqref{eq:mi}
		\State $I_s = CalculateMI(x_s)$;
		\State // \textit{Add into $\mathcal{D}$}
		\State $x_{best} \leftarrow x_{best} \cup x_s, \mathbf x \leftarrow \mathbf x \cup x_s$;
		\State $I_{best} \leftarrow I_{best} \cup I_s, \mathbf y \leftarrow \mathbf y \cup I_s$;		
		\EndIf
		\EndFor 
		\\
		\Return $x_{best}, I_{best}$
	\end{algorithmic} 
	\label{alg:bkiopt}
\end{algorithm}

\begin{proposition}
The time complexity of our proposed method at each while-loop step in Algorithm \ref{alg:bkiexp} is:
\begin{multline}
	\begin{matrix} \underbrace{ \mathcal{O}(N N_z N_c^2) } \\ \text{explicit MI evaluation} \end{matrix}
	+ \begin{matrix} \underbrace{ \mathcal{O} (N_{epoch}N\log{N_{q}}) } \\ \text{BKI MI inference} \end{matrix}
\end{multline}
where $N_{epoch}$ is the number of training epoch, $N_z$ and $N_c$ are the numbers of beams per sensor scan, and the number of cells that a beam intersects with the grid map at worst, respectively.
\end{proposition}

Significantly, the GP-based robot exploration in \cite{bai2015inference} and BO-based method in \cite{bai2016information} have the same time cost of ours in explicit MI evaluation, but these two methods have computational complexities of $\mathcal{O}(N^3+N^2 N_{q})$ and $\mathcal{O}(N_{epoch}(N^3+N^2 N_{q}))$ to perform the expensive GP inference for MI, respectively. This comparative theoretic result indicates our BKI-based exploration method outperforms the GP-based methods in time efficiency, especially in large-scale and cluttered places which need more samples $N$ and $N_{q}$ to evaluate rapidly.

\section{Results and Discussions}
In this section, we run numerical simulations and dataset experiments on a desktop PC with a 3.6 GHz Intel i3-9100F CPU and 32G RAM to verify the effectiveness of proposed BKI-based robot exploration method. The information threshold is $I_{th} = 0.05$ bit and the trade-off factor is $\alpha=0.5$. We adopt a Mat$\acute{\text e}$rn kernel for GP and the kernel parameters are $\ell=1$ and $\nu=3/2$ for all simulations. We also choose the parameters of $\zeta=0.001$ and $\sigma=0.01$ for BKI method. The robot poses are assumed to be known and the robot's candidate actions are sampled uniformly in the FOV of range sensors. We conduct 20 Monte Carlo trials for all maps. 

We use greedy-based optimization (named ``NBO'' in simulations), batch GP with only 1 epoch for optimization (``bacth GP'') \cite{bai2015inference}, and GP-based BO with multiple epochs (``GP-BO'') \cite{bai2016information} to compare with our methods, one named ``bacth BKI'' with only 1 optimization epoch and another one named ``BKI-BO'' with multiple epochs. Meanwhile, to validate the time efficiencies, we apply 2 cases of $N=30$ and $N=60$ samples for each method, where GP-BO 30 and BKI-BO 30 use $N_{epoch}=15$ iterations, GP-BO 60 and BKI-BO 60 use 30 epochs in BKI optimization. We also set $N_{q}=8N$ in all simulations.
\subsection{Synthetic Environments Results}
\begin{figure}
	\centering
	% 	\vspace{0.3em}
	\subfigure[Informative trajectory]{
		\begin{minipage}[t]{0.5\linewidth}
			\centering
			\includegraphics[width=1.8in]{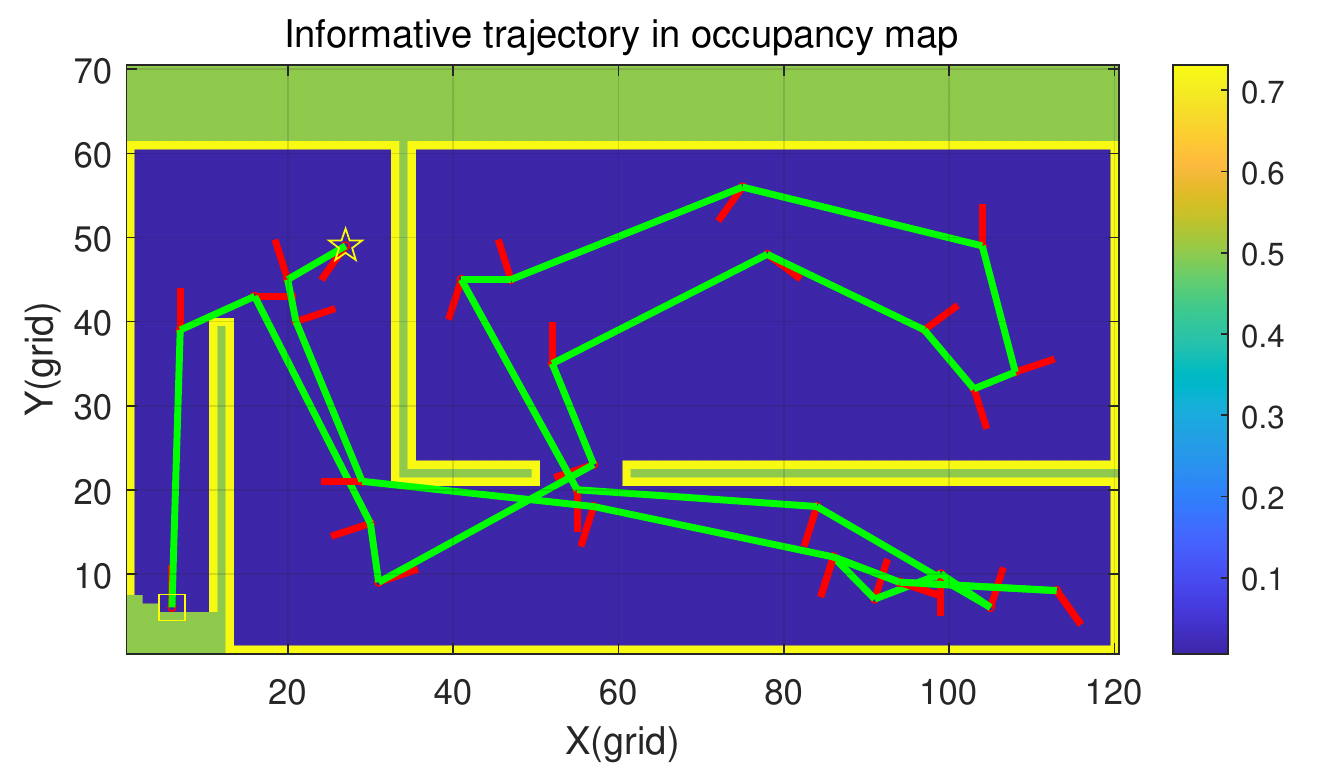}
			%\caption{fig1}
		\end{minipage}%
	}%
	% 	\quad
	\subfigure[MI surface]{
		\begin{minipage}[t]{0.5\linewidth}
			\centering
			\includegraphics[width=1.8in]{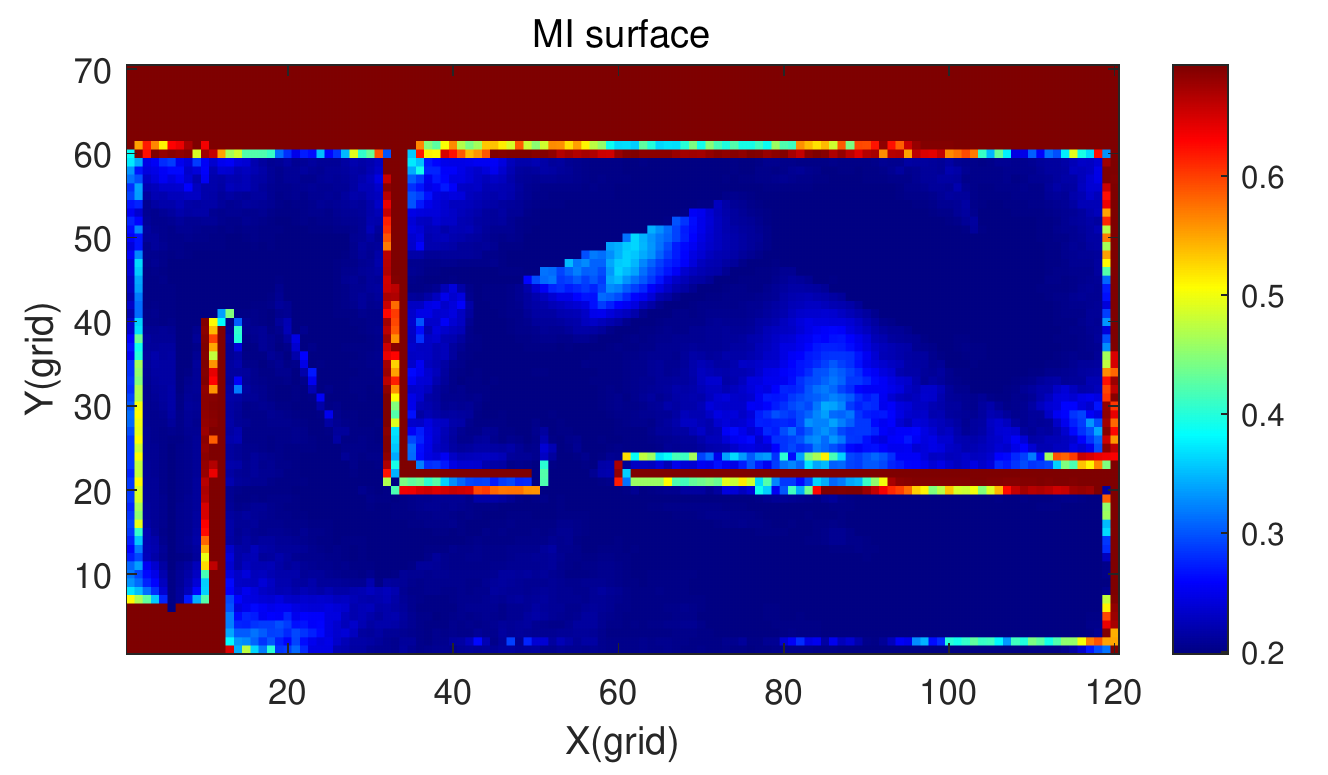}
			%\caption{fig2}
		\end{minipage}%
	}%
	% 	\vspace{0.3em}
	\centering
	\caption{An example of BKI-based robot exploration in an unknown \textbf{structured} environment. Yellow square: start point; yellow star: end point; red line: robot direction at each action.}
	\label{fig:str-traj}
\end{figure}

To simulate the indoor and field scenes, we generate 2 $24~\mathrm{m} \times 14 ~\mathrm{m}$ synthetic maps, one structured maze map surrounded by several walls (shown in Fig.~\ref{fig:str-traj}, $N_{loop}$ = 50), and one unstructured map consisting of circles and ellipses (shown in Fig.~\ref{fig:unstr-traj}, $N_{loop}$ = 150). The map resolutions are both 0.2 m. The simulated range sensor has a FOV of $\pm$1.5 rad with a resolution of 0.05 rad, and a maximum sensing range of $6$~m. The robot is initially at [1.2~m, 1.2~m] with 0~rad heading and trying to explore the prior unknown map. The representative resulting paths maximizing the information objective function are in Fig.~\ref{fig:unstr-traj} and Fig.~\ref{fig:str-traj}.

\begin{figure}[ht]
	\centering
	% 	\vspace{0.3em}
	\subfigure[Map entropy]{
		\begin{minipage}[t]{0.5\linewidth}
			\centering
			\includegraphics[width=1.8in]{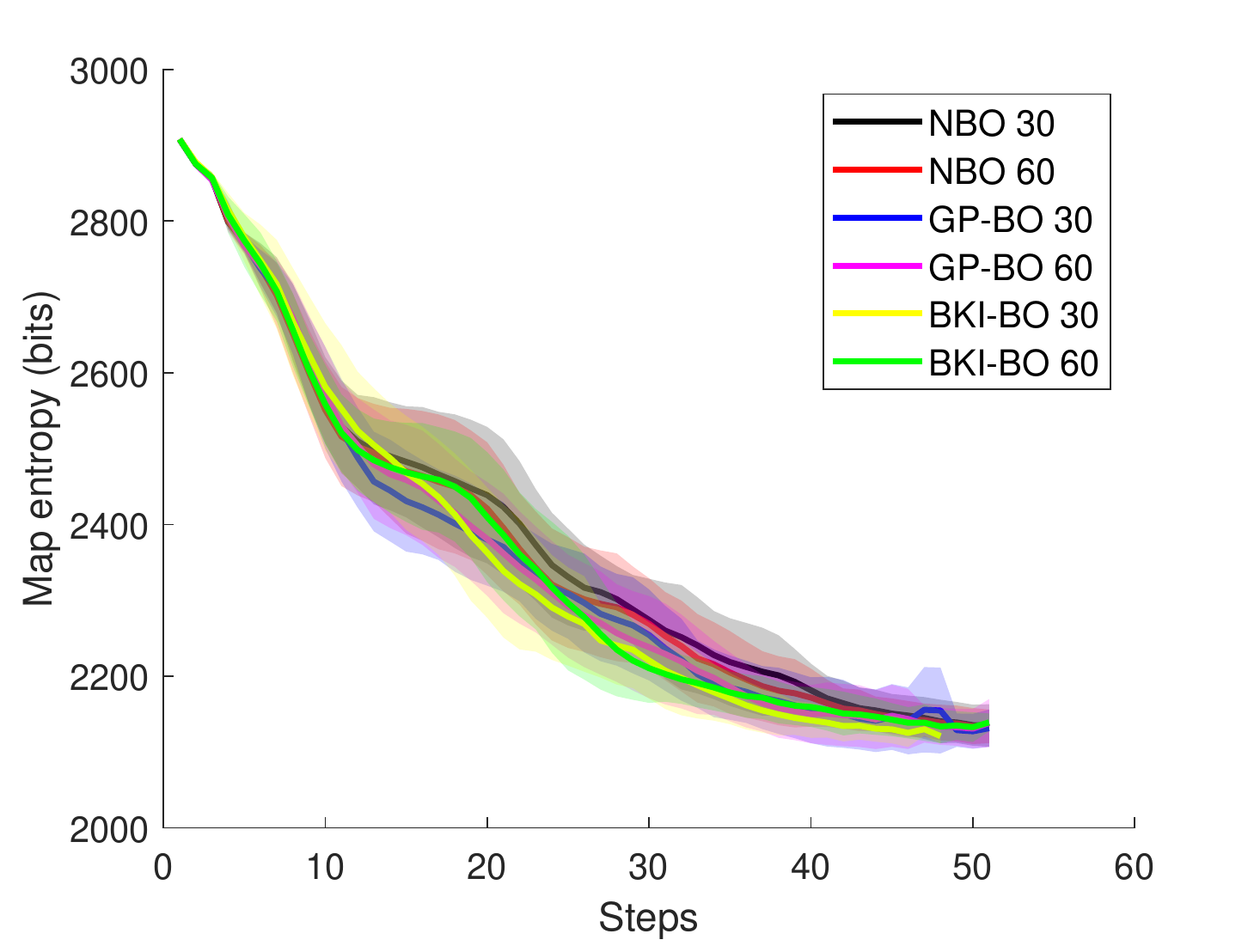}
			%\caption{fig1}
		\end{minipage}%
	}%
	\subfigure[Map entropy]{
		\begin{minipage}[t]{0.5\linewidth}
			\centering
			\includegraphics[width=1.8in]{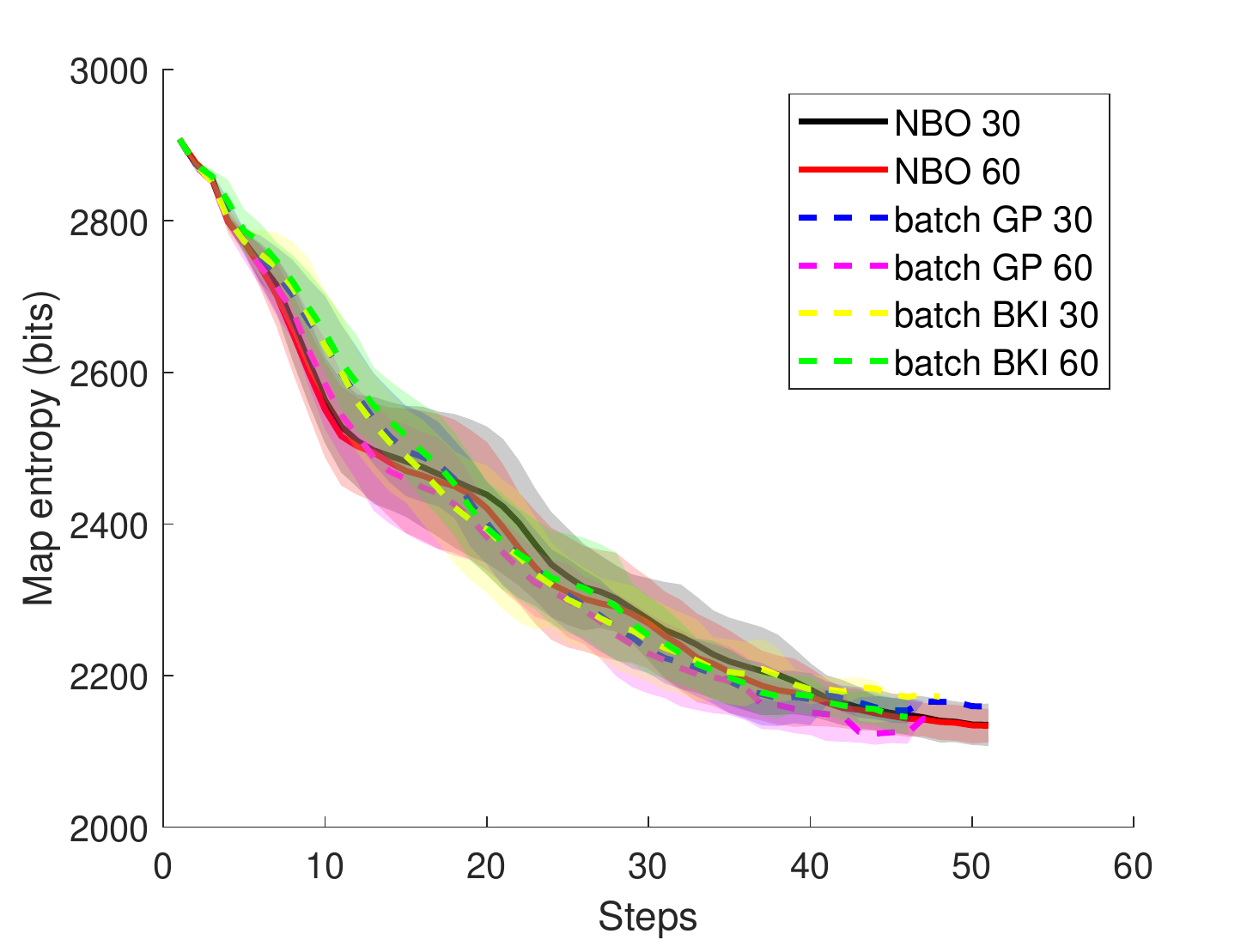}
			%\caption{fig2}
		\end{minipage}%
	}%
	\quad
	\subfigure[Coverage]{
		\begin{minipage}[t]{0.5\linewidth}
			\centering
			\includegraphics[width=1.8in]{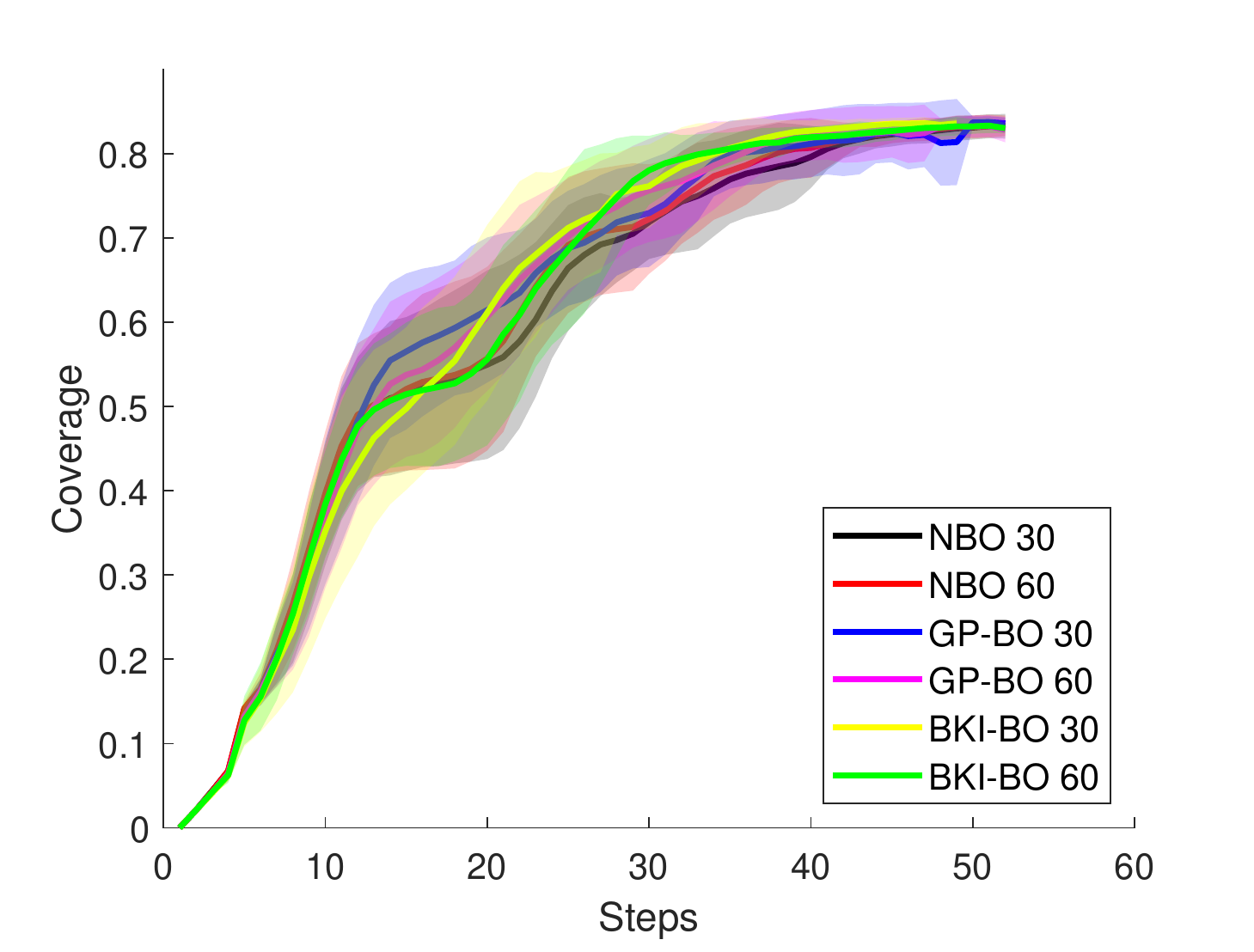}
			%\caption{fig1}
		\end{minipage}%
	}%
	\subfigure[Coverage]{
		\begin{minipage}[t]{0.5\linewidth}
			\centering
			\includegraphics[width=1.8in]{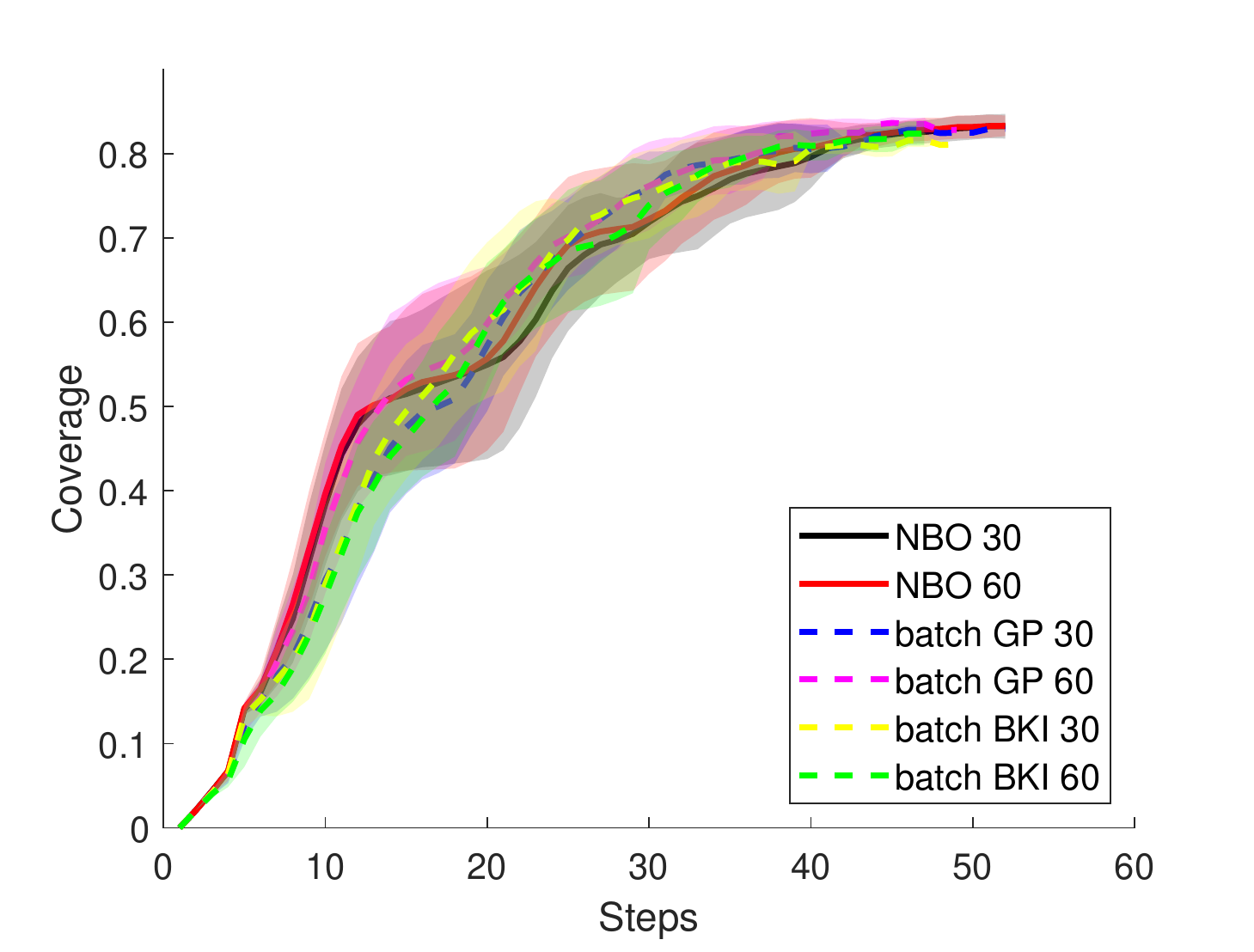}
			%\caption{fig2}
		\end{minipage}%
	}%
	\centering
	\caption{Map entropy and coverage results of the synthetic structured map.}
	\label{fig:str-res}
\end{figure}

\begin{figure*}[ht]
	\centering
	% 	\vspace{0.3em}
	\subfigure[]{
		\begin{minipage}[t]{0.25\linewidth}
			\centering
			\includegraphics[width=1.8in]{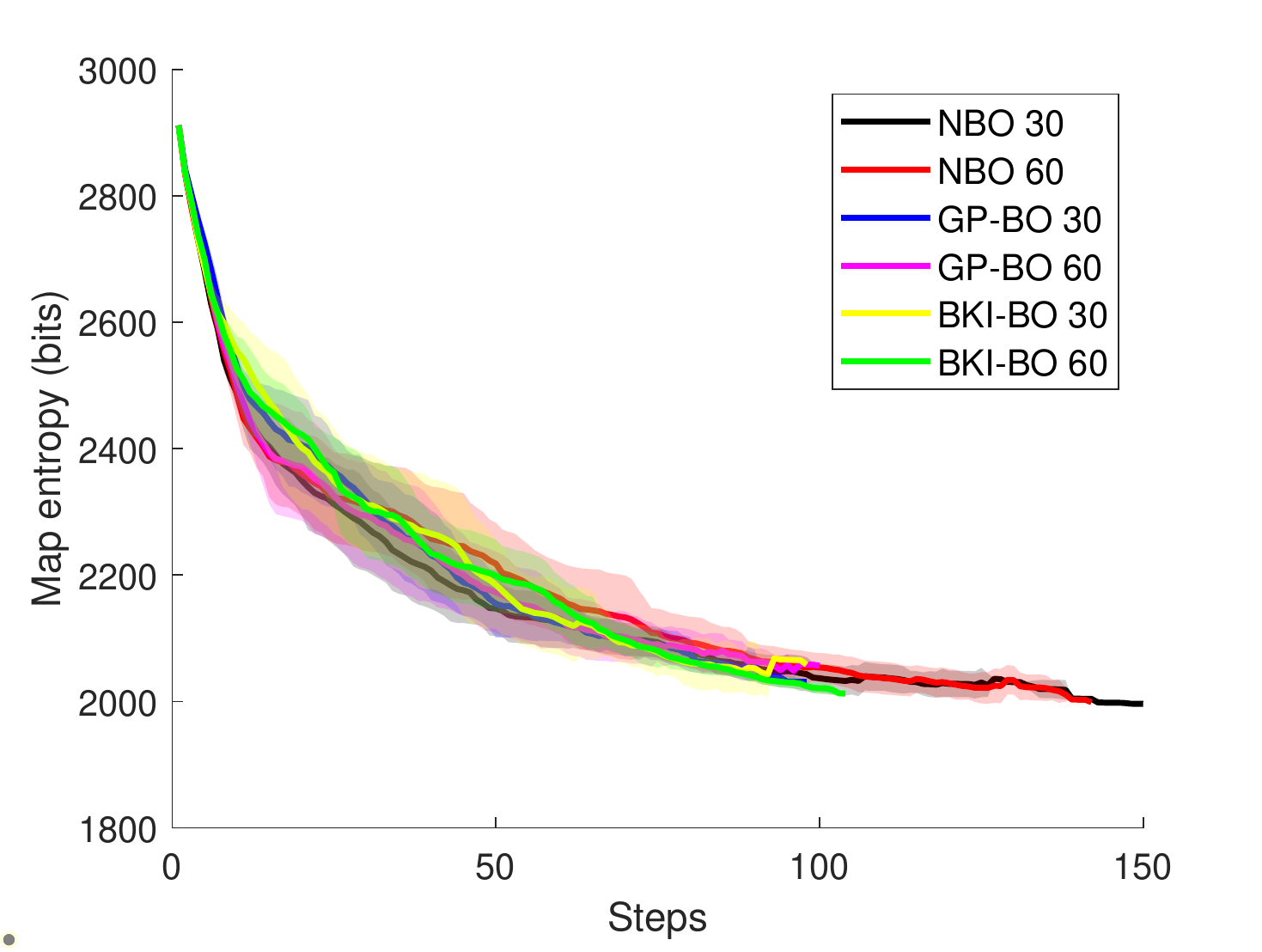}
			%\caption{fig1}
		\end{minipage}%
	}%
	\subfigure[]{
		\begin{minipage}[t]{0.25\linewidth}
			\centering
			\includegraphics[width=1.8in]{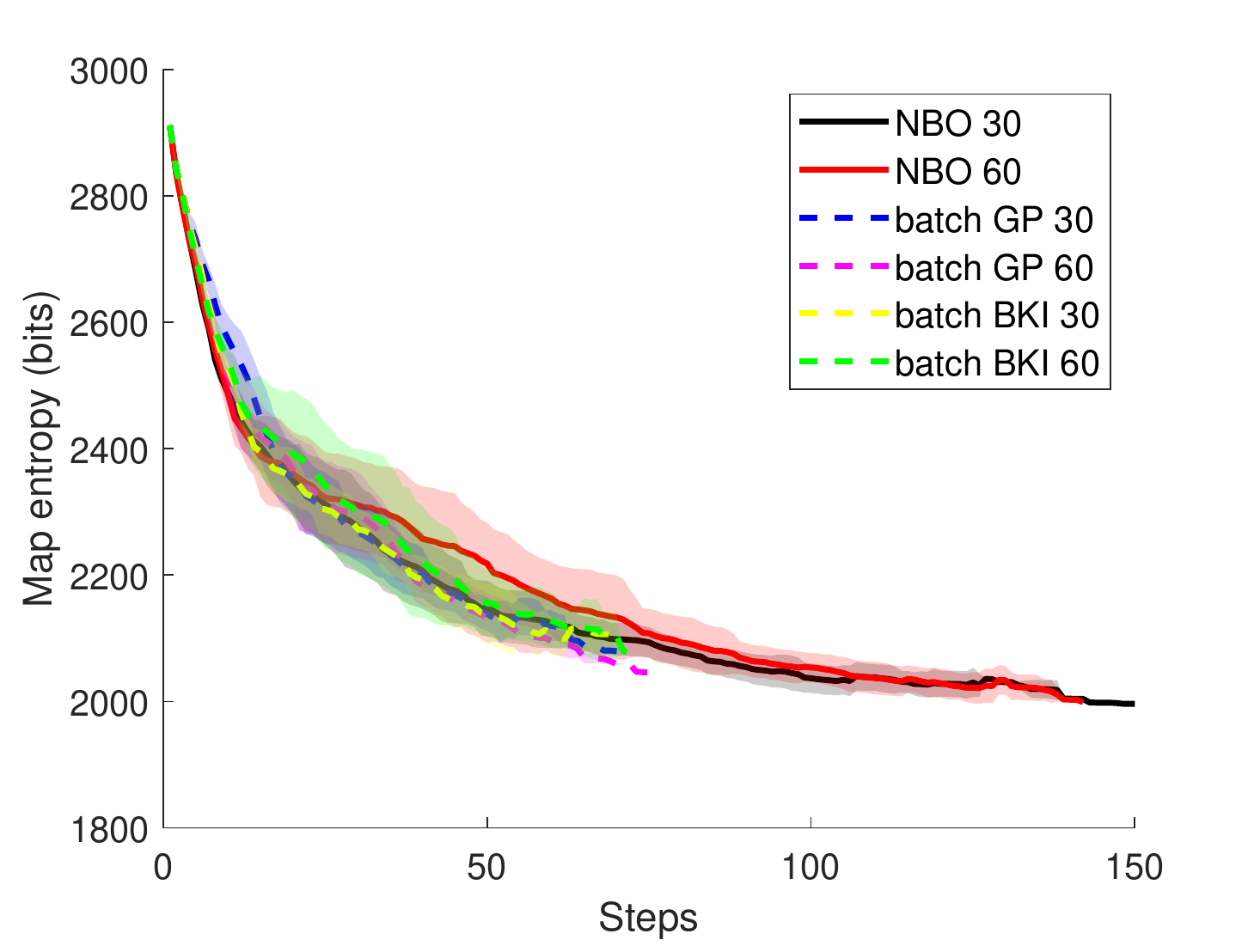}
			%\caption{fig2}
		\end{minipage}%
	}%
	% 	\quad
	\subfigure[]{
		\begin{minipage}[t]{0.25\linewidth}
			\centering
			\includegraphics[width=1.8in]{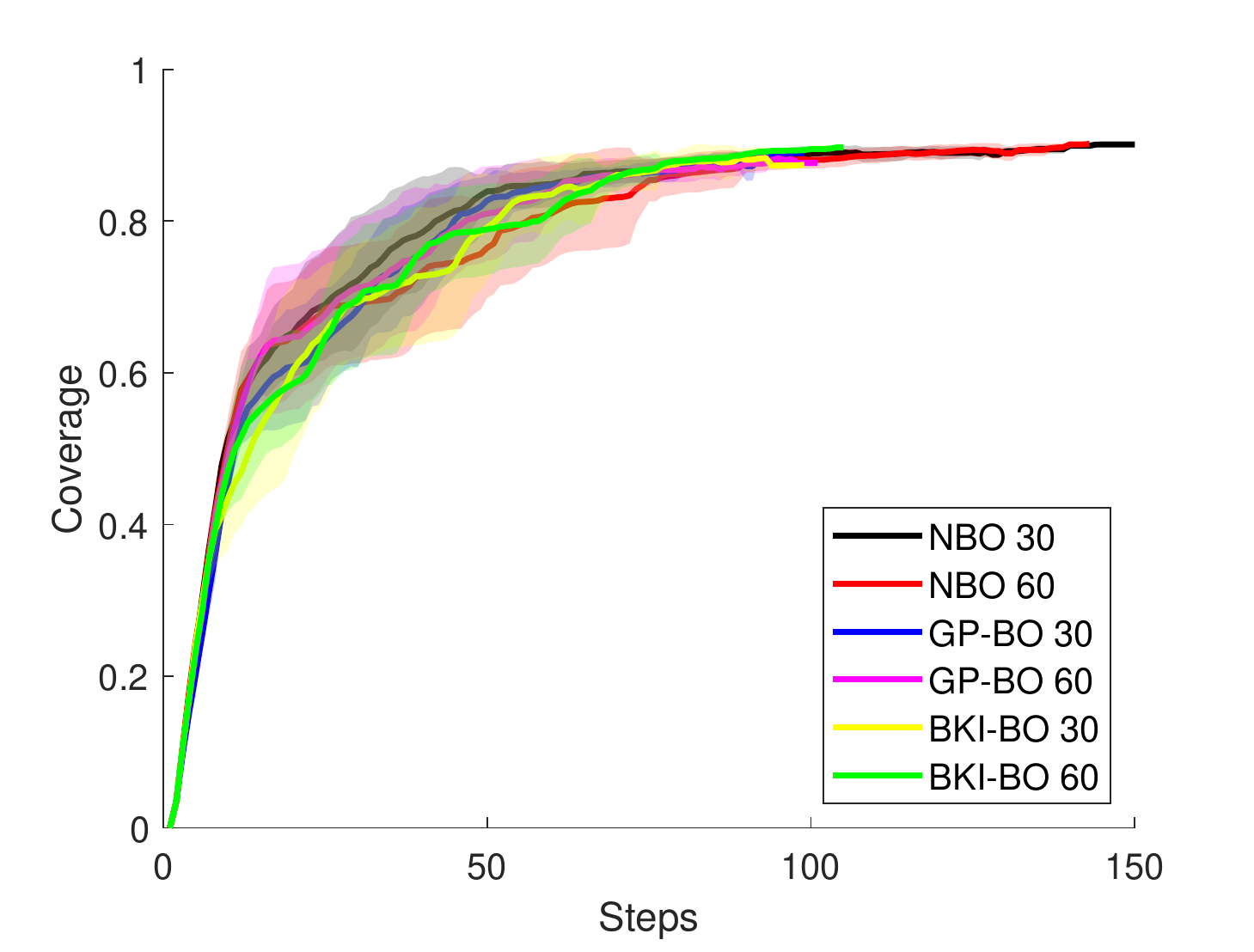}
			%\caption{fig1}
		\end{minipage}%
	}%
	\subfigure[]{
		\begin{minipage}[t]{0.25\linewidth}
			\centering
			\includegraphics[width=1.8in]{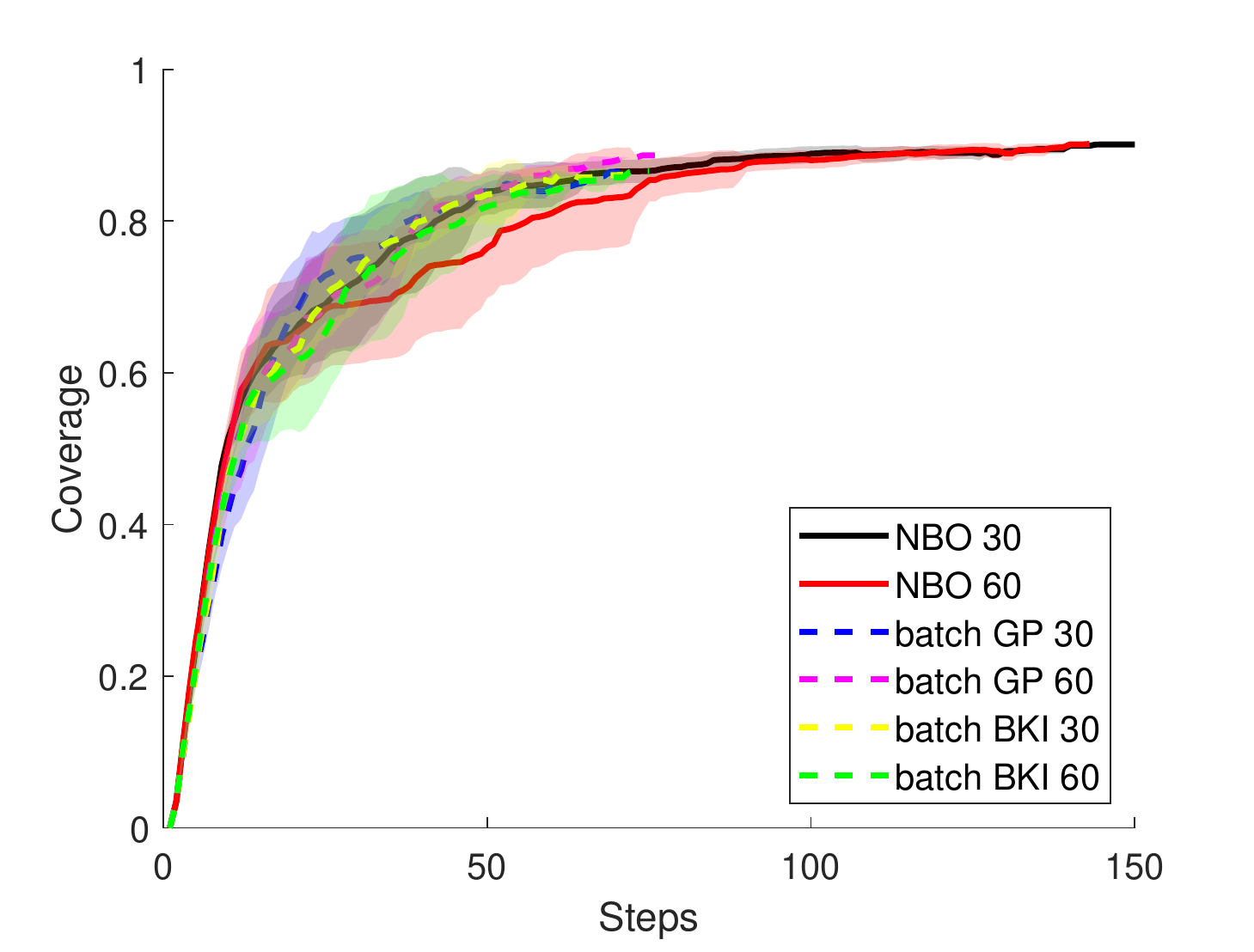}
			%\caption{fig2}
		\end{minipage}%
	}%
	\centering
	\caption{Map entropy and coverage results of the synthetic unstructured map.}
	\label{fig:unstr-res}
\end{figure*}

The qualitative results of structured and unstructured maps are shown in Fig.~\ref{fig:str-res} and Fig.~\ref{fig:unstr-res}, respectively. To compare the exploration performance using different methods intuitively, we present the evolution of map entropy and coverage rate of each method in the figures, where the solid and dashed lines depict the means of Monte Carlo trials for each method, and the shaded regions represent the standard deviations.

Fig.~\ref{fig:str-res} shows the BKI and GP methods have similar performance to the NBO methods since this structured scene is relatively small and simple, especially in the beginning stage where there is only one corridor to move forward. Differently,  Fig.~\ref{fig:unstr-res} indicates that the NBO methods spend more time (about 50$\sim$70 steps) to converge and end the exploration, while BKI and GP methods complete the exploration with comparable entropy reduction and coverage rates to NBOs.

\begin{figure}
	\includegraphics[width=3.0in]{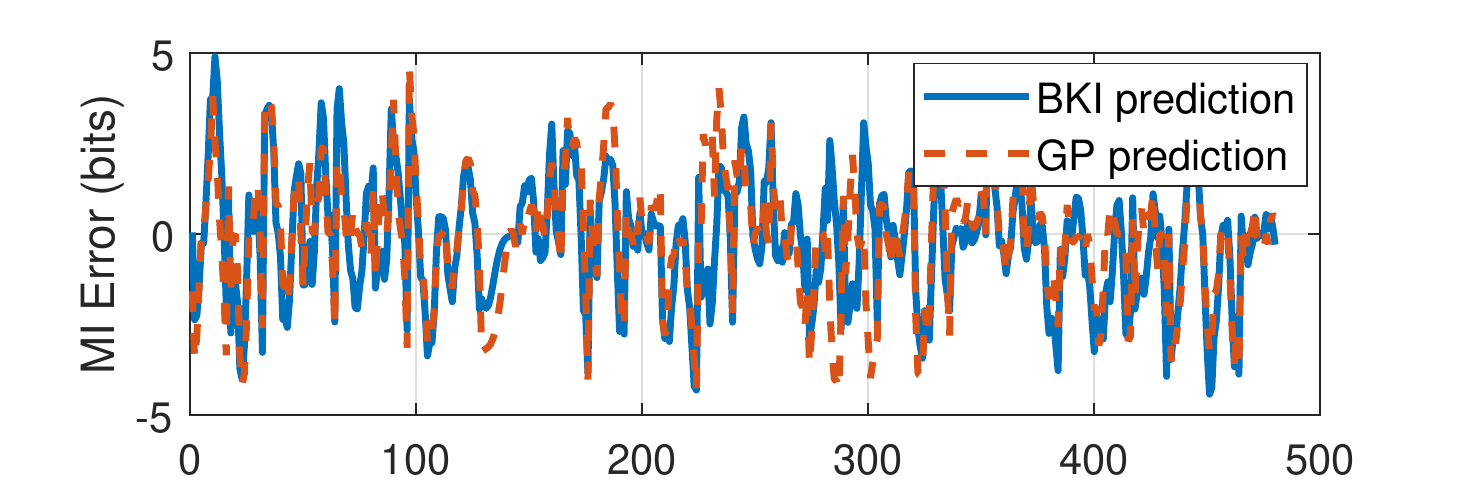}
	\centering
	\caption{A challenging example of MI prediction error comparison using BKI and GP methods trained with fewer samples in a randomly selected exploration step.}
	\label{fig:mi-error}
\end{figure}

Moreover, as in Fig.~\ref{fig:mi-error}, we use the explicitly evaluated MI as the ground truth and compute the MI prediction errors using BKI-BO and GP-BO methods with small training samples in a randomly selected step, which implies the BKI-based approach can resemble the GP-based one in MI inference accuracy when facing challenging cases.

In short, these results validate that our BKI methods have competitive properties with GP-based exploration ones in the typical structured and unstructured scenes.
%%%%%%%%%%%%%%%%%%%%%%%%%%%%%%%%%%%%%%%%%%%%%%%%%%%%%%%%%%%%%
%%%%%%%%%%%%%%%%%%%%%%%%%%%%%%%%%%%%%%%%%%%%%%%%%%%%%%%%%%%%%
%%%%%%%%%%%%%%%%%%%%%%%%%%%%%%%%%%%%%%%%%%%%%%%%%%%%%%%%%%%%
%%%%%%%%%%%%%%%%%%%%%%%%%%%%%%%%%%%%%%%%%%%%%%%%%%%%%%%%%%%%

%%%%%%%%%%%%%%%%%%%%%%%%%%%%%%%%%%%%%%%%%%%%%%%%%%%%%%%%%%%%
%%%%%%%%%%%%%%%%%%%%%%%%%%%%%%%%%%%%%%%%%%%%%%%%%%%%%%%%%%%%
%%%%%%%%%%%%%%%%%%%%%%%%%%%%%%%%%%%%%%%%%%%%%%%%%%%%%%%%%%%%%
%%%%%%%%%%%%%%%%%%%%%%%%%%%%%%%%%%%%%%%%%%%%%%%%%%%%%%%%%%%%%
\subsection{Dataset Results}
\begin{figure}
	\centering
	% 	\vspace{0.3em}
	\subfigure[Exploration trajectory]{
		\begin{minipage}[t]{0.5\linewidth}
			\centering
			\includegraphics[width=1.8in]{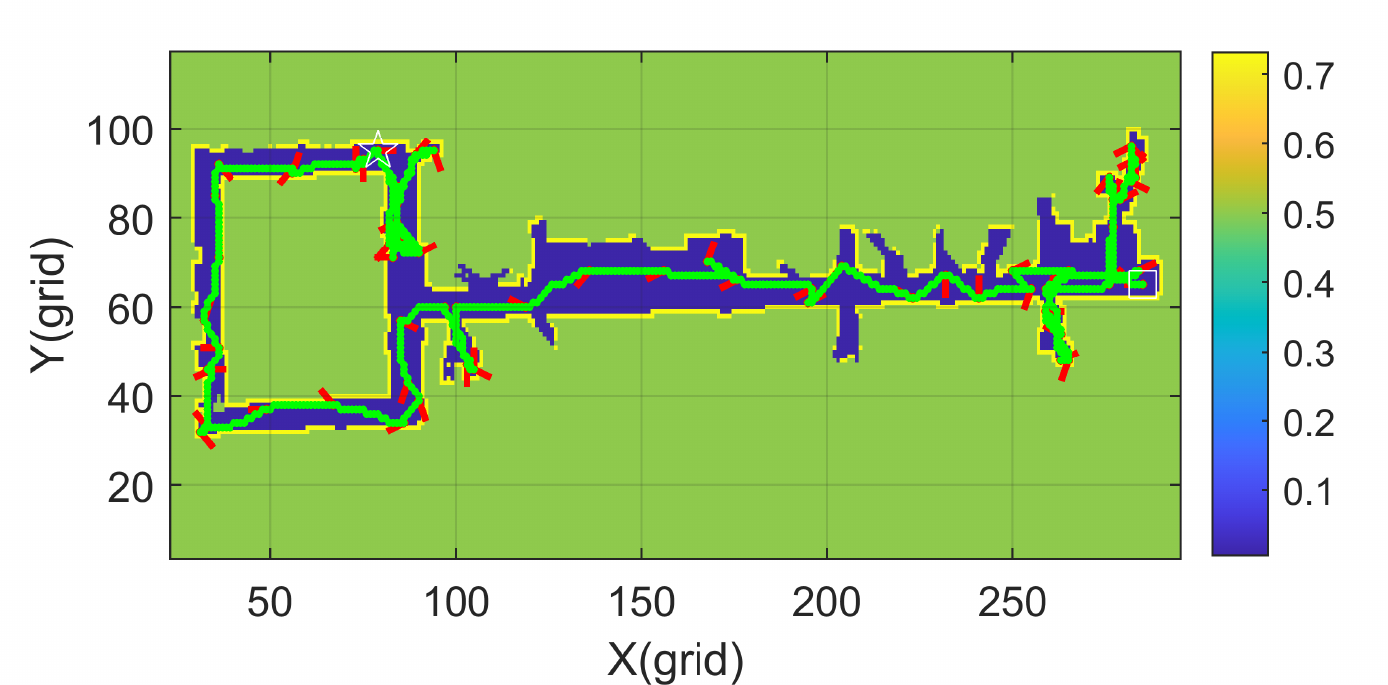}
			%\caption{fig1}
		\end{minipage}%
	}%
	% 	\quad
	\subfigure[MI surface]{
		\begin{minipage}[t]{0.5\linewidth}
			\centering
			\includegraphics[width=1.8in]{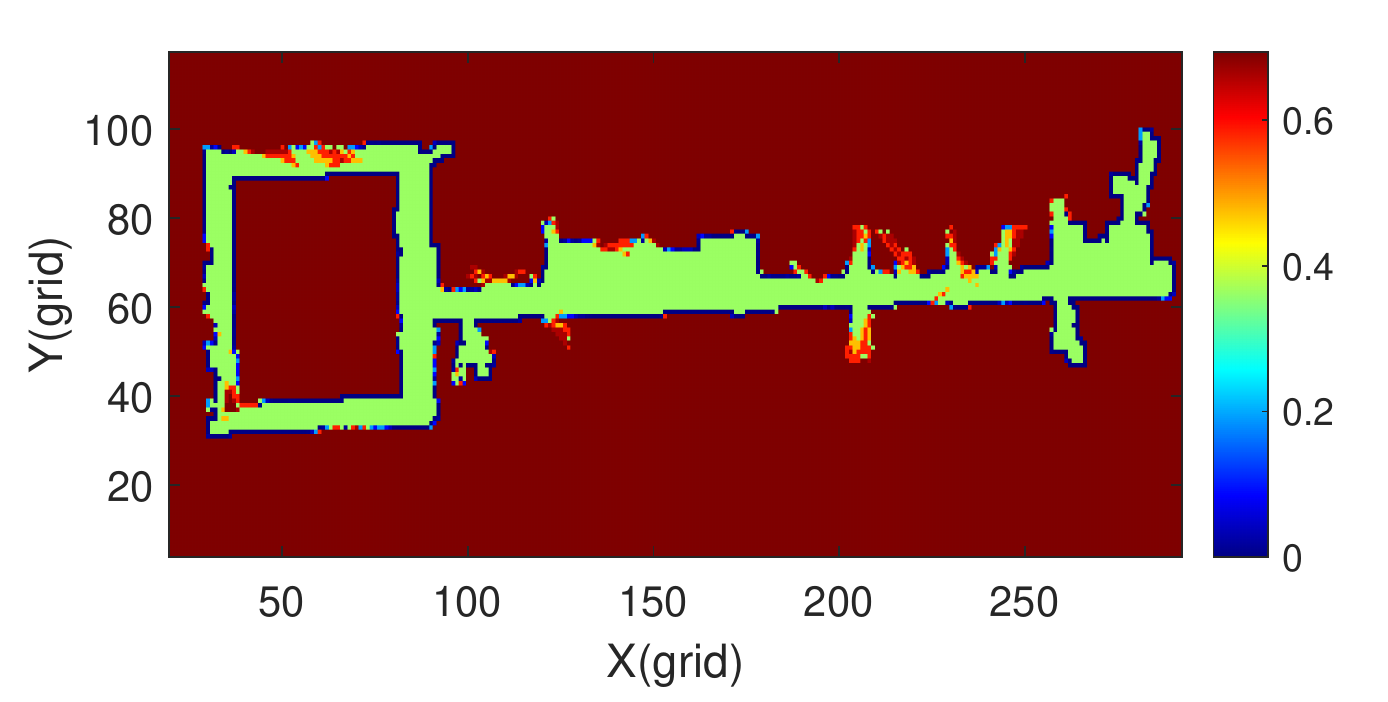}
			%\caption{fig2}
		\end{minipage}%
	}%
	% 	\vspace{0.3em}
	\centering
	\caption{An example of BKI-based robot exploration in the large cluttered Seattle map \cite{Radish}. White square: start point; White star: end point.}
	\label{fig:seattle-traj}
\end{figure}

To test our method in a more complex environment, we choose the Seattle map \cite{Radish} containing narrow long corridors and cluttered rooms, as in Fig.~\ref{fig:seattle-traj}. The map size is $24~\mathrm{m} \times 14 ~\mathrm{m}$ with a resolution of 0.2 m. We use a simulated laser scanner emitting 20 beams uniformly within a FOV of $\pm \pi/3$ rad at a maximum range of 4 m. The robot starts at [13, 57]m with a $-\pi/2$ initial heading angle. The $N_{loop}$ is set to 100. 

\begin{figure}[ht]
	\centering
% 	\vspace{0.3em}
	\subfigure[Map entropy rate]{
		\begin{minipage}[t]{0.5\linewidth}
			\centering
			\includegraphics[width=1.8in]{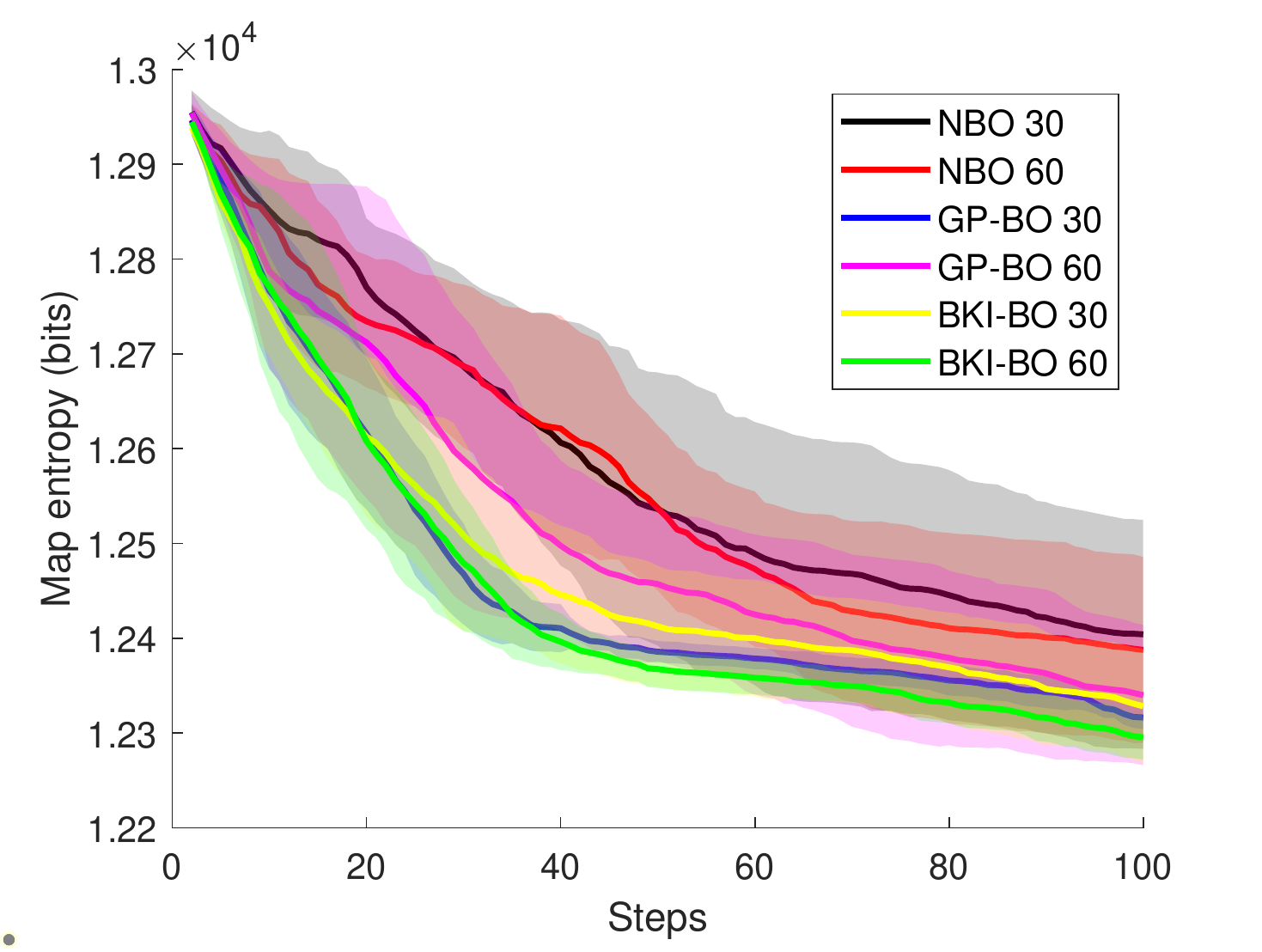}
			%\caption{fig1}
		\end{minipage}%
	}%
	\subfigure[Coverage rate]{
		\begin{minipage}[t]{0.5\linewidth}
			\centering
			\includegraphics[width=1.8in]{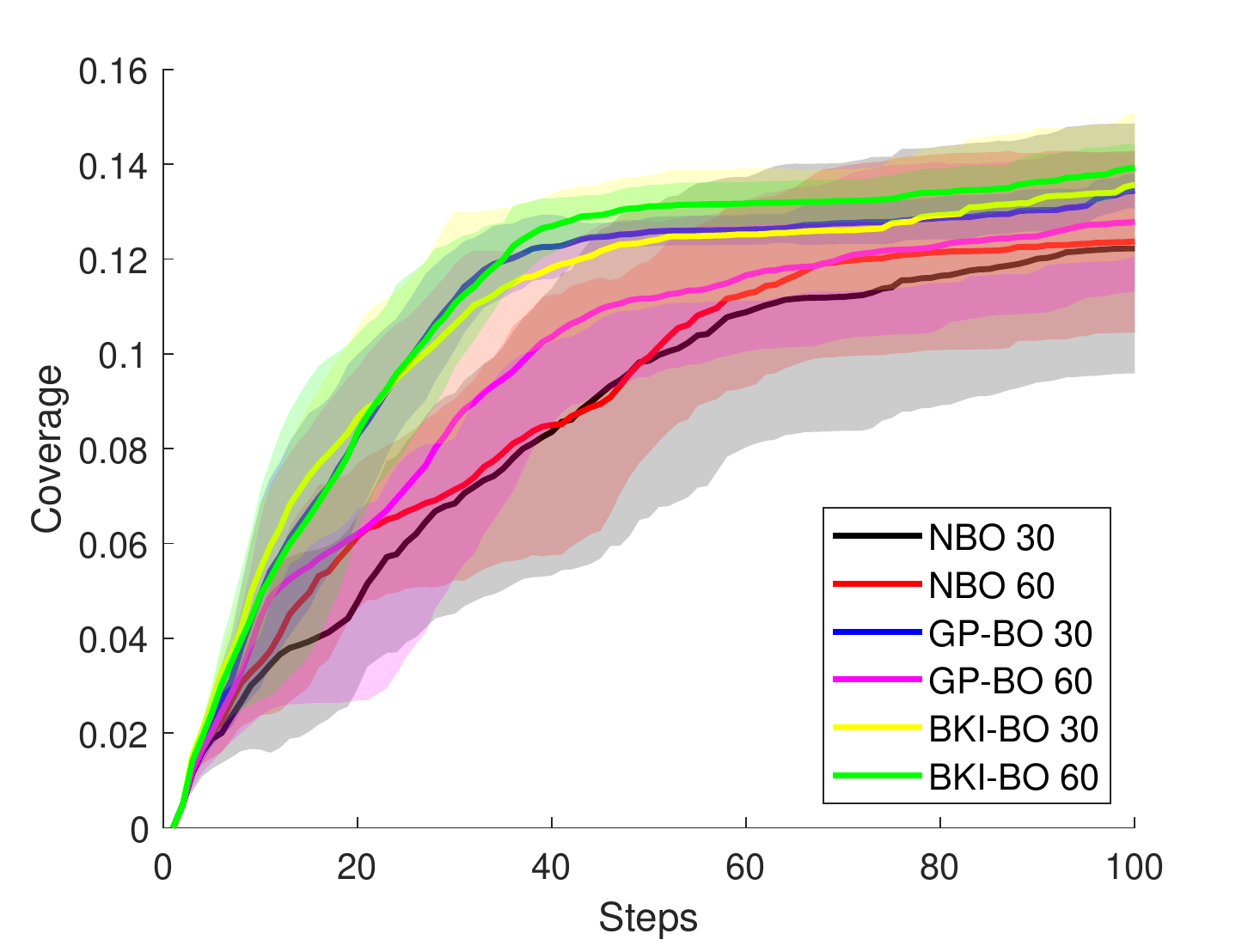}
			%\caption{fig1}
		\end{minipage}%
	}%
	\centering
	\caption{Map entropy and coverage results of the Seattle map results (batch methods omitted). }
	\label{fig:sea-res}
\end{figure}

Fig.~\ref{fig:sea-res} presents the comparative curves of map entropy and coverage rates, whereas Fig.~\ref{fig:sea-res}(a) shows the BKI-BO methods have more rapid reduction rates of map entropy after the exploration starts and arrive at relatively lower levels than other methods, among them, BKI-BO 60 performs the best. In this typical cluttered map, GP-BO methods perform slightly inferior to our BKI-BO methods but almost catch up with ours, which also are much better than the NBO methods. 
The curves in Fig.~\ref{fig:sea-res}(b) imply that batch GP and batch BKI have similar performance.
We also can get an insight from Fig.~\ref{fig:sea-res}(c) and (d), i.e., the coverage curves of BKI-BO methods converge slightly earlier than GP-BO methods and reach higher values, and all BO-based methods explore the unknown place much faster than the NBO ones. This result evidences our BKI methods are more suitable for large cluttered environments.
%%%%%%%%%%%%%%%%%%%%%%%%%%%%%%%%%%%%%%%%%%%%%%%%%%%%%%%%%%%%%
%%%%%%%%%%%%%%%%%%%%%%%%%%%%%%%%%%%%%%%%%%%%%%%%%%%%%%%%%%%%%
%%%%%%%%%%%%%%%%%%%%%%%%%%%%%%%%%%%%%%%%%%%%%%%%%%%%%%%%%%%%
%%%%%%%%%%%%%%%%%%%%%%%%%%%%%%%%%%%%%%%%%%%%%%%%%%%%%%%%%%%%
%%%%%%%%%%%%%%%%%%%%%%%%%%%%%%%%%%%%%%%%%%%%%%%%%%%%%%%%%%%%%
\begin{table*}
	\caption{Time cost comparison of different exploration methods }
	\begin{center}
		\begin{tabular}{c c c c}
			\hline
			Methods & Synthetic structured map & Synthetic unstructured map & Seattle map \cite{Radish}\\ \hline
			NBO 30  & 95.29\%~/~10.4455 $\pm$ 0.9409 & 96\%~/~12.1683 $\pm$ 1.3856 & 96.95\%~/~4.8434 $\pm$ 0.7311\\ 
			NBO 60  & 95.38\%~/~10.9967 $\pm$ 1.0676 & 95.93\%~/~12.4971 $\pm$ 2.1583 & 96.93\%~/~5.3502 $\pm$ 0.9009\\ 
			batch GP 30 & 5.15\%~/~0.4387 $\pm$ 0.0246 & 4.66\%~/~0.2805 $\pm$ 0.0232 &12.68\%~/~0.2134 $\pm$ 0.0169\\ 
			batch GP 60 & 6.44\%~/~0.4444 $\pm$ 0.0487 & 5.89\%~/~0.3021 $\pm$ 0.0362 & 14.94\%~/~0.2291 $\pm$ 0.0226\\ 
			batch BKI 30 (ours) & 3.05~/~0.4324 $\pm$ 0.0276 & 2.93\%~/~0.2485 $\pm$ 0.0346 & 7.67\%~/~0.2036 $\pm$ 0.0254 \\ 
			batch BKI 60 (ours) & 3.87\%~/~0.4407 $\pm$ 0.0384 & 3.56\%~/~0.2731 $\pm$ 0.0356 & 9.05\%~/~0.2065 $\pm$ 0.0229\\ 
			GP-BO 30 & 49.71\%~/~0.9435 $\pm$ 0.0609 & 48.09\%~/~0.6083 $\pm$ 0.1121 & 67.97\%~/~0.5203 $\pm$ 0.0554\\ 
			GP-BO 60 & \textbf{74.03\%~/~1.8265 $\pm$ 0.1189} & \textbf{72.99\%~/~1.3558 $\pm$ 0.1190} & \textbf{84.26\%~/~1.0528 $\pm$ 0.1124}\\ 			 
			BKI-BO 30 (ours) & 39\%~/~0.7518 $\pm$ 0.0683 & 39.14\%~/~0.514 $\pm$ 0.0966 30 & 54.03\%~/~0.3903 $\pm$ 0.1175\\ 
			\textbf{BKI-BO 60 (ours)} & \textbf{53.74\%}~/~\textbf{0.9952 $\pm$ 0.1061} & \textbf{54.31\%~/~0.7363 $\pm$ 0.1186} & \textbf{62.45\%}~/~\textbf{0.4955 $\pm$ 0.1775}\\ \hline
			\multicolumn{4}{l}{Note: Time cost of inference per step (in percentage) / Total time cost of exploration per step of each method (in sec.)}
		\end{tabular}
		\label{tab:time}
	\end{center}
	% 	\vspace{-0.5cm}
\end{table*}
%%%%%%%%%%%%%%%%%%%%%%%%%%%%%%%%%%%%%%%%%%%%%%%%%%%%%%%%%%%%%
\subsection{Time Efficiency}
We have presented the exploration results in the previous simulations of typical scenes, and our BKI-based method has shown desired exploration performance in efficiency and accuracy compared with state-of-the-art methods. 
To put more intuitive and specific comparison, we further analyze the time cost of each method per exploration step in all maps. As in Table~\ref{tab:time}, the results show the time cost of the whole exploration process per step in the form of means and standard deviations, as well as the average percent of evaluation and decision-making time spent by different methods in each step.

Among the 10 methods, the basic NBO methods have the most expensive time consumption (more than about 8$\sim$50 times to BKI and GP methods) per step, while other methods based on GP and BKI cost much less time, showing the efficiency of Bayesian optimization-based approaches. We can further analyze these results from 2 aspects of view. From the top row to the bottom, our BKI-based methods get better time efficiency performance of decision-making and inference than the corresponding GP-based ones in all maps when the number of samples increases, e.g. batch BKI 30/60 vs batch GP 30/60 and BKI-BO 30/60 vs GP-BO 30/60. We also can observe that BKI methods run faster than GP ones when using more training epochs. BKI methods also bring significant time savings for exploration, such as decreasing by about 20\% and 45\% time compared with GP-BO 30 and GP-BO 60 respectively in the structured map. 

From the left column to the right, these above-mentioned differences get more distinct in unstructured and large cluttered maps, e.g. the time costs per step of GP-BO 30 and GP-BO 60 decrease by about 25\% and 53\% in the Seattle map respectively, which also verifies our proposed BKI-based robot exploration methods can improve the time efficiency considerably without losing overall exploration performance compared with other methods. 

\section{Conclusions}

This paper mainly contributed to a new efficient learning-based approach for information-theoretic robot exploration in unknown environments. In particular, a continuous information gain evaluation model for predicting the MI of numerous sampled robot actions is built by introducing the Bayesian kernel inference method. The time complexity of MI prediction is decreased to logarithm level in comparison with state-of-the-art methods. An objective function integrating the predicted MI and uncertainty is also designed to balance exploration and exploitation. 
The proposed method also gets verified under an autonomous exploration framework by extensive simulations of different scenes, which reveals our method outperforms the greedy-based and GP-based exploration methods overall in efficiency without loss of exploration performance, especially in unstructured and large cluttered scenes. 
Future work mainly involves studying the exploration performance using different $\alpha$ values and kernels, as well as extending our method to 3D scenes.


%%%%%%%%%%%%%%%%%%%%%%%%%%%%%%%%%%%%%%%%%%%%%%%%%%%%%%%%%%%%%%%%%%%%%%%%%%%%%%%%

% \section*{ACKNOWLEDGMENT}

%%%%%%%%%%%%%%%%%%%%%%%%%%%%%%%%%%%%%%%%%%%%%%%%%%%%%%%%%%%%%%%%%%%%%%%%%%%%%%%%
\bibliographystyle{IEEEtran}
\bibliography{icra23.bib}

\end{document}